\documentclass{article}
\usepackage{spconf,amsmath,epsfig}
\usepackage{amsmath, amsfonts, booktabs, multirow}
\let\OLDthebibliography\thebibliography
\renewcommand\thebibliography[1]{
  \OLDthebibliography{#1}
  \setlength{\parskip}{0pt}
  \setlength{\itemsep}{0pt plus 0.3ex}
}

\begin{document}\sloppy

\def\x{{\mathbf x}}
\def\L{{\cal L}}

\title{Towards Escaping from  Language Bias and OCR Error:  \\Semantics-Centered Text Visual Question Answering}
%
\name{Chengyang Fang\textsuperscript{1,2}, Gangyan Zeng\textsuperscript{3}, Yu Zhou\textsuperscript{1,2,*},  Daiqing Wu \textsuperscript{1,2},Can Ma \textsuperscript{1,2,*},Dayong Hu \textsuperscript{4}, Weiping Wang\textsuperscript{1}
     }
      
\address{\textsuperscript{1}Institute of Information Engineering, Chinese Academy of Sciences, Beijing, China  \\
        \textsuperscript{2}School of Cyber Security, University of Chinese Academy of Sciences, Beijing, China  \\
        \textsuperscript{3}School of Information and Communication Engineering, Communication University of China, Beijing, China  \\
        \textsuperscript{4}Heilongjiang Network Space Research Center, Harbin 150010, China  \\
        {\{fangchengyang, zhouyu, wudaiqing, macan, wangweiping\}@iie.ac.cn},\\ zgy1997@cuc.edu.cn,superhudayong@163.com}

\maketitle

\begin{abstract}
Texts in scene images convey critical information for scene understanding and reasoning. 
The abilities of reading and reasoning matter for the model in the text-based visual question answering (TextVQA) process.
However, current TextVQA models do not center on the text and suffer from several limitations.
The model is easily dominated by language biases and optical character recognition (OCR) errors due to the absence of semantic guidance in the answer prediction process.
In this paper,  we propose a novel Semantics-Centered Network (SC-Net) that consists of an instance-level contrastive semantic prediction module  (ICSP)  and a semantics-centered transformer module (SCT).
Equipped with the two modules, the semantics-centered model can resist the language biases and the accumulated errors from OCR. 
Extensive experiments on TextVQA and ST-VQA datasets show the effectiveness of our model. SC-Net surpasses previous works with a noticeable margin and is more reasonable for the TextVQA task.
\end{abstract}
\section{Introduction}
\label{sec:intro}
Visual-impaired people usually ask questions that require reading texts in the scene,  e.g.,  ``what year is the wine?''. As a semantically rich entity,  text plays a critical role in scene understanding and reasoning. However, most existing visual question answering (VQA) datasets \cite{goyal2017making} and methods \cite{anderson2018bottom, jiang2018pythia} mainly focus on the visual components and ignore the texts in images. To address this issue,  the text-based visual question answering (TextVQA) task \cite{singh2019towards} is proposed to endow  VQA models  the ability to read and analyze scene texts.

\begin{figure}[h]
  \centering
  \includegraphics[width=1\linewidth]{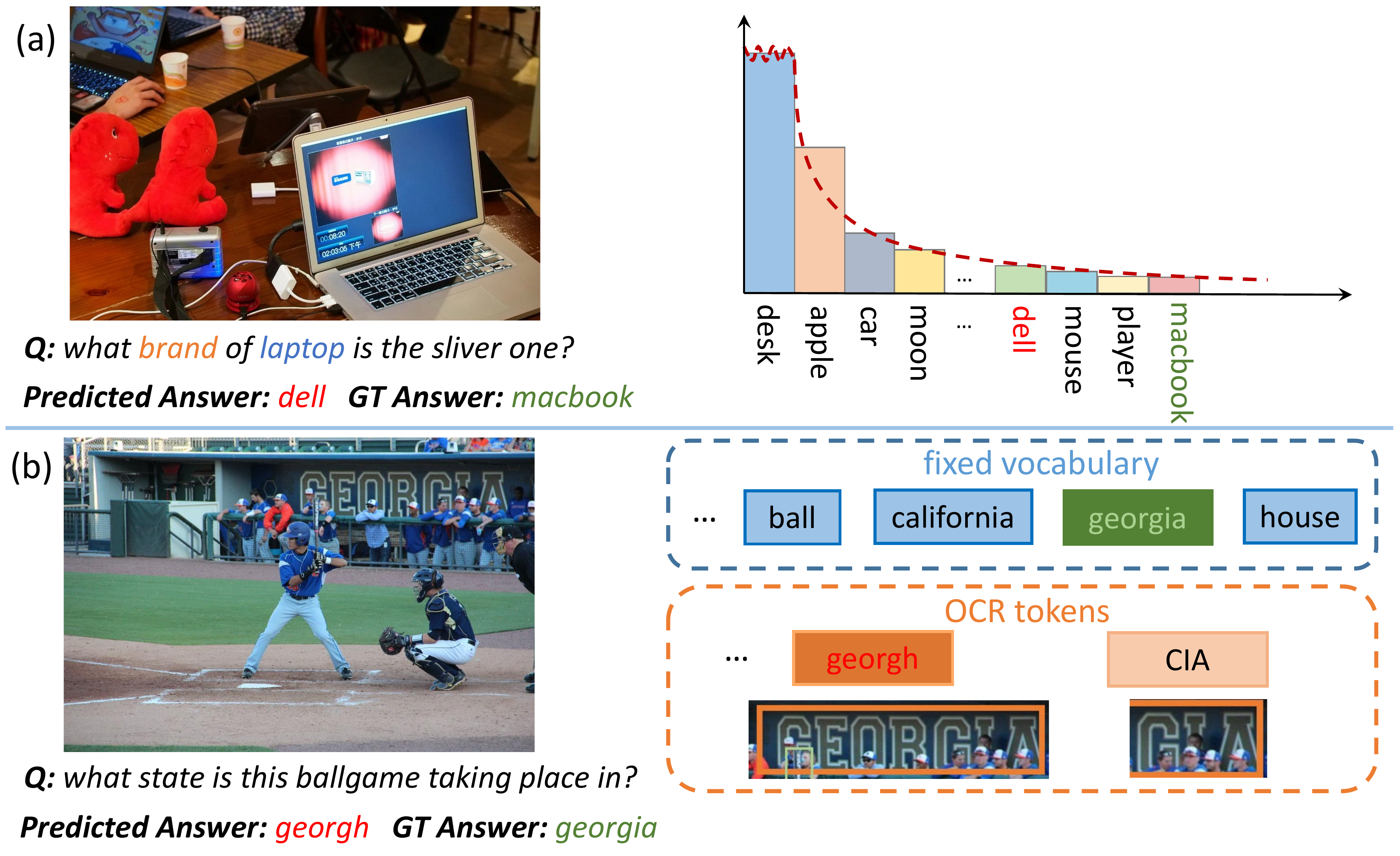}
  \caption{Illustration of the challenges in TextVQA. (a) TextVQA models tend to select the word that occurs frequently with the question information as the answer, failing to find the correct answer. (b) TextVQA models are easily affected by the OCR accuracy since they simply copy the OCR result without considering the semantic rationality.}
  \label{figure_1}
\end{figure}

LoRRA \cite{singh2019towards} extends previous VQA models \cite{anderson2018bottom, jiang2018pythia} with an external optical character recognition (OCR)  module and uses a copy mechanism    to  build a dynamic vocabulary containing OCR tokens.
M4C \cite{hu2020iterative} enriches OCR representations and utilizes a multimodal transformer \cite{vaswani2017attention} to project all entities from each modality (questions, OCR tokens and visual objects) to a joint embedding space.
Although notable improvements have been achieved with recent promising techniques \cite{liu2020cascade,han2020finding,zhu2020simple,yang2021tap}, many challenges still exist.

Generally, the answers to the text-based questions are sequentially selected from a fixed vocabulary that consists of high-frequency words in training set and a dynamic vocabulary composed of OCR result tokens in the image. The answer decoding in each iteration is usually modeled as a classification problem on the concatenation of $M$ high-frequency words and $N$ OCR tokens. In order to cover more candidate answer words, the pre-defined fixed vocabulary is usually quite large. For example, $M$ is set to 5,000 on TextVQA dataset \cite{singh2019towards} in M4C. However, the search space of $M$ plus $N$ is so large that the model has difficulty finding the correct words, especially when the training data are limited. We find the answer in the fixed vocabulary has a long tail distribution, and the words at the tail are often hard to learn. As shown in Figure \ref{figure_1}(a), the model simply selects the word ``dell'' as the answer as it frequently occurs together with the question words ``brand'' and ``laptop''.

On the other hand, for the dynamic vocabulary which contains OCR tokens, the accuracy of the OCR-copying answer heavily depends on the performance of the OCR module. For example, in Figure \ref{figure_1} (b), when the OCR module detects the answer text (``georgia") but has a  error recognition  (``georgh"), the model still tends to copy the wrong recognition result even the correct answer exists in the fixed vocabulary.

In fact, TextVQA is genuinely a more semantics-oriented task compared with general VQA and text spotting tasks. However, previous works do not put enough emphasis on the semantics. Therefore, the models cannot distinguish the answer text from a semantic perspective, resulting in the problems mentioned above.
Intuitively, we think introducing semantic supervision on decoding progress and focusing on semantic information in the pipeline can  alleviate these problems. Therefore, we propose a novel Semantics-Centered Network (SC- Net) that consists of an instance-level contrastive semantic prediction module (ICSP) and a semantics-centered trans- former module (SCT). 

First, current TextVQA models 
ignore the semantic rationality between the predicted answer and the information of the input. To tackle this drawback, we design an ICSP to explicitly involve semantic information to guide the answer decoding. 
To acquire a meaningful answer's semantic information, we develop an instance-level contrastive learning method to make the predicted semantic feature closer to the ground-truth answer’s feature while farther away from the non-answer words’ features. 

Second, the semantic information extracted by existing multimodal fusion modules is interfered by many other information. Specifically, current TextVQA models simply concatenate all modality entities (questions, OCR tokens and visual objects) into the interaction module (e.g. transformer) and treat them indiscriminately. There are two kinds of limitations: 1) semantic information is not the focus and 2) the input length is too long which causes large computation costs. 
To solve this problem, we propose a SCT which focuses on the semantic information in the multimodal fusion stage.

In summary, this paper's contributions are as follows:
\begin{itemize}
  \vspace{-0.75em}
  \item [1)  ] We propose a semantics-centered TextVQA framework which puts more emphasis on semantics compared with previous works since TextVQA is more semantics-related than the general VQA task.
  \vspace{-0.75em}
  \item [2)  ] An instance-level contrastive semantic prediction module (ICSP) to decode answers with semantic guidance is designed. 
  It eliminates the negative effects of language biases and can leverage the fixed vocabulary to compensate for OCR errors.
  \vspace{-0.75em}
  \item [3)  ] A semantics-centered transformer module (SCT) is proposed to highlight semantics in multimodal fusion and reduce computation costs.
  \vspace{-0.75em}
  \item [4)  ] Extensive experiments on several benchmarks demonstrate  our proposed method is able to provide a more semantically reasonable answer and outperforms several SOTA approaches in both accuracy and ANLS metrics.
  \end{itemize}

\begin{figure*}[h]
  \centering
  \includegraphics[width=1\linewidth]{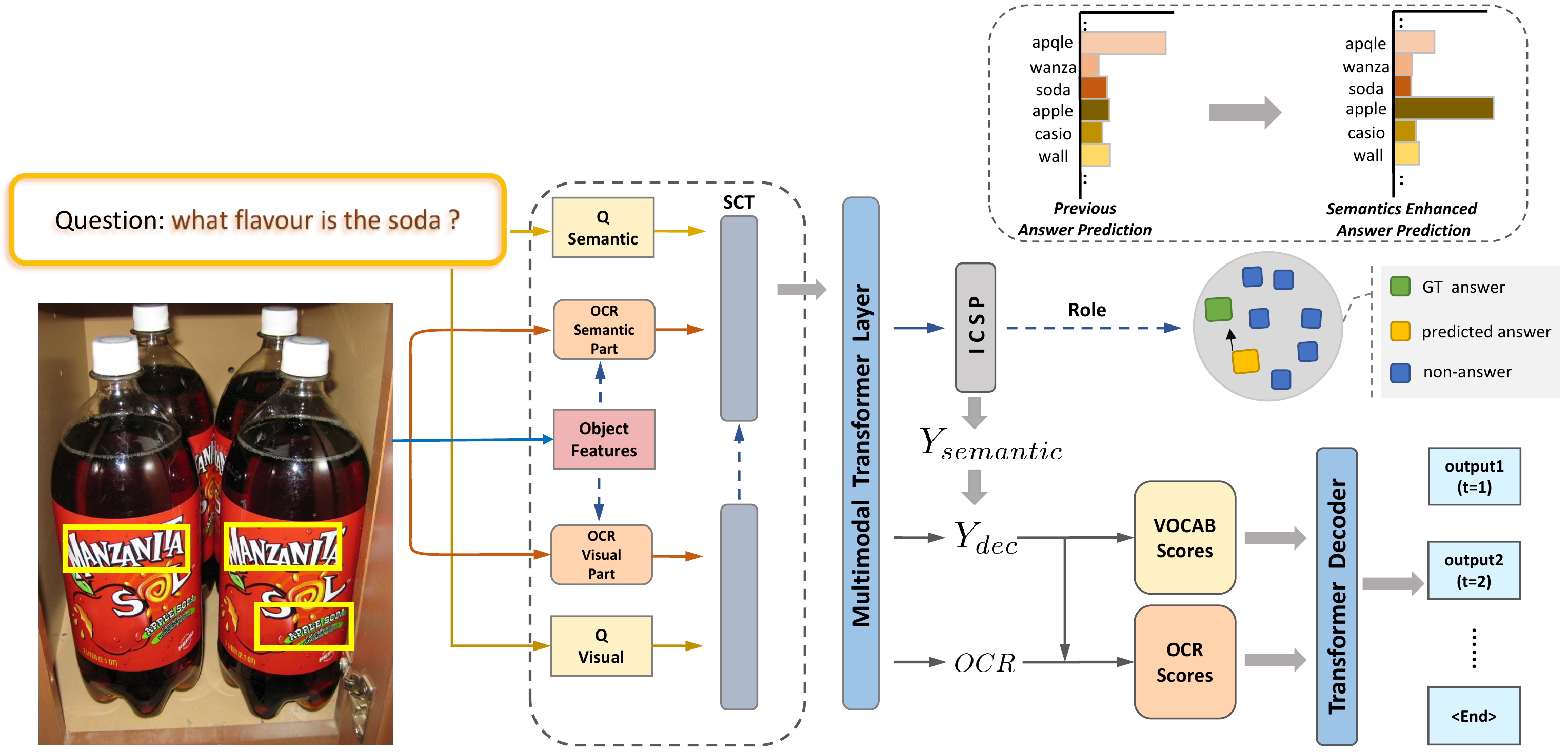}
  \caption{An overview of the semantics-centered network  (SC-Net). 
  }
  \label{figure_2}
\end{figure*}

\section{Related Work}
\subsection{Text-based Visual Question Answering} 
TextVQA attracts more and more attention from communities since it focuses on texts in natural daily scenes, such as road signs and displays. To promote progress in this field, several datasets \cite{singh2019towards,biten2019scene,mishra2019ocr} and methods\cite{han2020finding,singh2019towards,wang2020general,hu2020iterative} have been proposed. LoRRA \cite{singh2019towards} is the pioneering model which utilizes an attention mechanism to handle image features,  OCR features,  and question words, but it can only select one word as the answer. The Multimodal Multi-Copy Mesh   (M4C)   model \cite{hu2020iterative} boosts TextVQA performance by employing a multimodal transformer \cite{vaswani2017attention} to fuse various modality entities. 
In the following works, modifications with respect to feature embedding \cite{zhu2020simple}, feature interaction \cite{gao2020structured, liu2020cascade, kant2020spatially} and answer decoding\cite{han2020finding}   have been shown to improve performance. In addition, a  pre-training method TAP   \cite{yang2021tap} is proposed on the TextVQA and TextCaption tasks. It's an orthogonal technique to above methods.  So, we do not include it in our comparison.


\vspace{-1em}
\subsection{Semantics Enhanced Answer Prediction } 
In the recent work, some efforts are dedicated to generating a semantically driven answer in TextVQA. Evidence-based VQA \cite{wang2020general} has been proposed to use IoU to indicate the evidence to guide the scene reasoning.  LaAP-Net \cite{han2020finding} extends it by designing a localization-aware answer predictor which generates a bounding box in the image as evidence of the answer. The supplementary evidence helps determine whether the generated answer has been reasonably inferred. In contrast, our work avoids the multi-step evidence generation process and directly introduces semantic information to guide answer generation.

\vspace{-1em}
\section{Method}

In this paper, we focus on answering text-based visual questions in a semantically driven manner. The pipeline of our model is shown in Figure \ref{figure_2}.\\
\textbf{Question Features.} Following M4C  \cite{hu2020iterative},  we use a three-layer BERT  \cite{devlin2018bert} model to obtain the question feature embedding \begin{math}
  Q=\left \{{q_{i}}\right \}_{i=1}^{L}
\end{math},  where   \begin{math}
  q_{i}\in \mathbb{R}^{d}
\end{math}
is the $i$-th question word embedding,  $L$ is the length of the question,  and $d$ is the dimension of the feature. For aligning with visual and semantic features of texts separately,  we design two individual self-attention layers \cite{vaswani2017attention} to obtain question feature embedding \begin{math}
  Q_{v} 
\end{math} for visual-part branch and \begin{math}
  Q_{s} 
\end{math} for semantic-part branch.\\ 
\textbf{OCR Features.} For each text recognized by the OCR system in the input image,  we encode four different features and combine them in a complementary way to visual and semantic parts.
\begin{align}
\begin{split}
     x_{i}^{ocr, v}=&LN  ( W_{fr}x_{i}^{ocr, fr} )  +LN  ( W_{bx}x_{i}^{ocr, bx} )   \\
  x_{i}^{ocr, s}=&LN  ( W_{ft}x_{i}^{ocr, ft}+W_{ph}x_{i}^{ocr, ph} )  +LN  ( W_{bx}x_{i}^{ocr, bx}  ) \\
   x_{i}^{ocr, iou}=& W_{iou}IoU ( x_{i}^{ocr, bx} ) 
  \end{split}
\end{align}where \begin{math}x_{i}^{ocr, v}\end{math} is the OCR visual-part feature and \begin{math}x_{i}^{ocr, s}\end{math} is the OCR semantic-part feature.  The OCR visual-part features consist of 1) appearance feature \begin{math}x_{i}^{ocr, fr}\end{math} extracted from the fc6 layer of a pretrained Faster R-CNN \cite{ren2015faster} model,  and 2) the bounding box feature \begin{math}x_{i}^{ocr, bx} = [\frac{x_{i}^{min}}{W}, \frac{y_{i}^{min}}{H}, \frac{x_{i}^{max}}{W}, \frac{y_{i}^{max}}{H}]\end{math}. The OCR semantic-part features are made up of 1)  FastText \cite{bojanowski2017enriching} feature \begin{math}x_{i}^{ocr, ft}\end{math} ,  2)  PHOC \cite{almazan2014word} feature \begin{math}x_{i}^{ocr, ph}\end{math} ,  and 3)  the bounding box feature \begin{math}x_{i}^{ocr, bx} \end{math}. $LN$ is Layer Normalization  and  $IoU$ is a function to calculate the IoU between \begin{math}x_{i}^{ocr, bx} \end{math} and other text and object regions in the image. \\
\textbf{Object Features.} To get object region features,  we apply the same Faster R-CNN model mentioned in OCR visual-part features. 
\begin{align}
\begin{split}
      x_{i}^{obj}=&LN  ( W_{fr}^{'}x_{i}^{obj, fr} )  +LN  ( W_{bx}^{'}x_{i}^{obj, bx} )   \\
      x_{i}^{obj, iou}=& W_{iou}^{'}IoU ( x_{i}^{obj, bx} ) 
\end{split}
\end{align}
where \begin{math}x_{i}^{obj, fr}\end{math} is the appearance feature and \begin{math}x_{i}^{obj, bx}\end{math} is the bounding box feature.  $IoU $is a function to calculate the IoU between \begin{math}x_{i}^{obj, bx}\end{math} and other text and object regions in the image.

\vspace{-1em}
\subsection{Semantics-Centered Transformer }
Existing works \cite{liu2020cascade, hu2020iterative, kant2020spatially, singh2019towards}
ignores the different roles of objects and texts in the TextVQA task. 
To better exploit the information beneficial for reasoning, we model the semantic features specifically and augment them with the aid of visual features. The object features are used to enhance the visual and semantic text features adaptively. The proposed SCT module is shown in Figure 3. The inputs of the SCT are formulated as:
\begin{align}
\begin{split}
     x^{ocr, v_e}&= att ( x^{ocr, iou}, x^{obj, iou}, x^{obj} )  + x^{ocr, v} \\
     x^{ocr, s_e}&= att ( x^{ocr, iou}, x^{obj, iou}, x^{obj} )  + x^{ocr, s}\\
     input^{v}&=[Q_{v}, x^{ocr, v_e}]\\
     input^{s}&=[Q_{s}, x^{ocr, s_e}]
     \end{split}
\end{align}
where [ , ] means concatenation  operation and $att$ means attention mechanism.

\begin{figure}[h]
  \centering
  \includegraphics[width=0.8\linewidth]{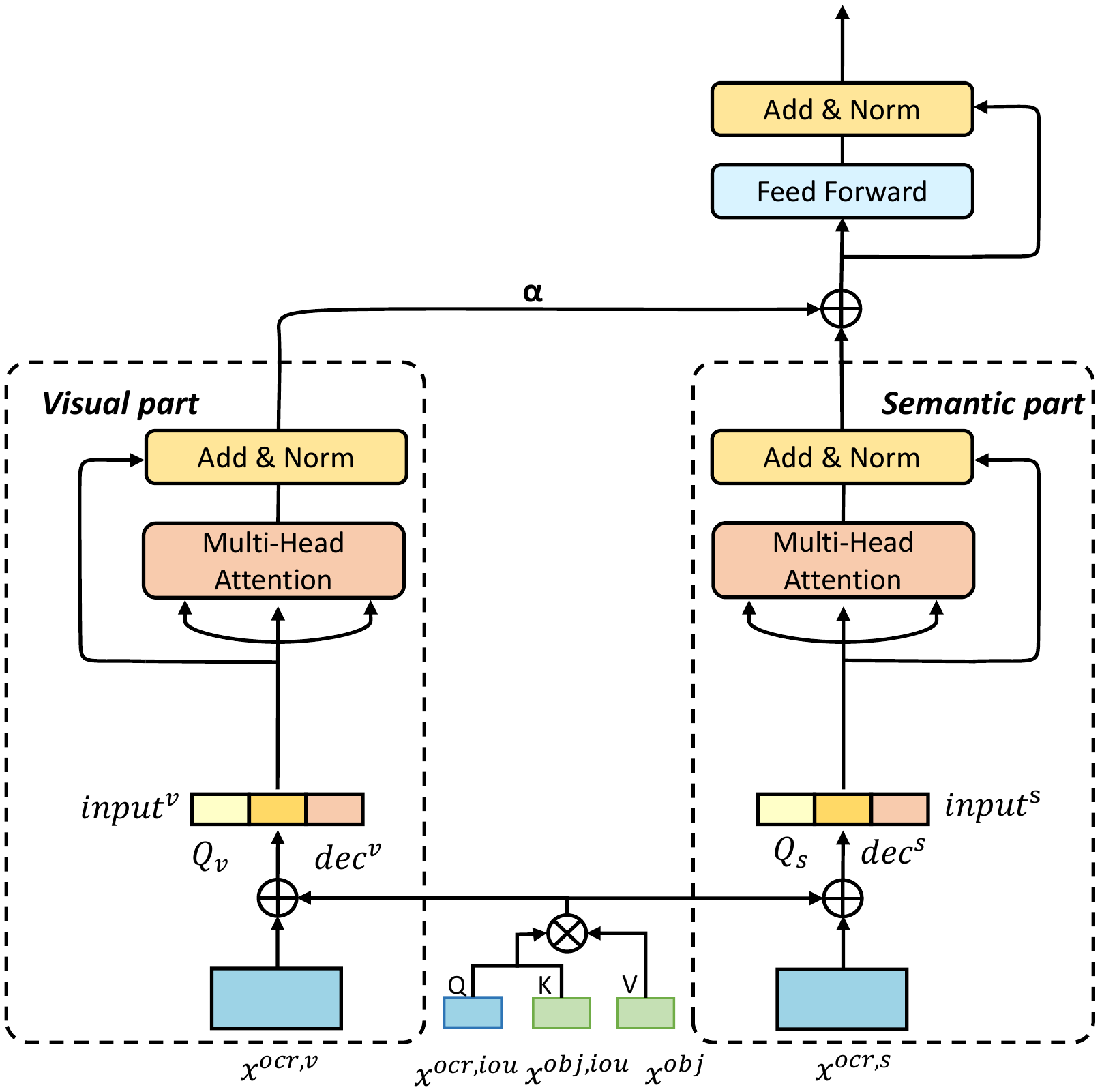}
  \caption{Architecture of semantics-centered transformer. 
  }
  \label{figure_3}
\end{figure}

As we can see in Figure \ref{figure_3}, the visual and semantic features are modeled individually with a two-branch architecture. Because IOU between indicate the spatial relation between regions, we use the IoU information of  OCR tokens and objects as query and key respectively in the attention mechanism to integrate objects with OCR tokens because it contains spatial information of regions. 
Since semantic information accounts for more contribution,  the visual-part output  acts as assistant information to improve network performance. We fuse them as: 
\begin{align}
\begin{split}
    output^{s}&=att ( input^{s}, input^{s}, input^{s} ) \\
    output^{v}&=att ( input^{v}, input^{v}, input^{v} ) \\
    z^{fuse} &= output^{s} + \alpha * output^{v}
    \end{split}
\end{align} where \begin{math} output^{s}\end{math} and \begin{math} output^{v}\end{math} denote outputs of the semantic branch and the visual branch respectively. Note that the \begin{math}
  \alpha
\end{math} is a learnable parameter in training process. In this manner,  the model can adaptively determine the proportion of contributions of semantic and visual features so as to fully exploit the advantages of multimodal information. 
The output \begin{math}
   z^{fuse}
\end{math} goes through the feed-forward layer of the semantic branch as the output of SCT .

\subsection{Instance-Level Contrastive Semantic Prediction }
Due to the quite large classification answer space in TextVQA, 
if there is no external information guidance or tremendous training data, TextVQA models are easily affected by language biases. We propose the ICSP module to alleviate this problem via predicting the global semantic embedding of the answer in advance, and combining it with the original decoding embedding to guide answer generation. 
 
 Aiming to predict the semantic information of the answer,  we use the output of the transformer corresponding to a special token \begin{math}
  [CLS]
\end{math} as the input of ICSP,  because it contains the global information of input embedding features. We utilize a two-layer MLP to project the \begin{math}
  [CLS]^{output}
\end{math} to the same dimension as the FastText embedding of the answer as follow:
\begin{equation}
    ans^{se} = W_{1}*  ( W_{2}*[CLS]^{output}+bias_{2} )  +bias_{1}
\end{equation}
where \begin{math}
   W_{1}\text{,   } W_{2} \text{,   } bias_{1}\text{,   } bias_{2}
\end{math} are weights and biases of the MLP. Then the predicted answer semantic information is embedded and added to the decoder output to generate the final semantically enhanced answer representation \begin{math}
   z^{ans}
\end{math} as
\begin{equation}
    z^{ans} = y^{dec} + \alpha^{se} \circ   ( W^{se}*ans^{se}+bias^{se} )  
\end{equation} where \begin{math}
  \alpha^{se} \text{ is a learnable ratio parameter}
\end{math}. For OCR-copying scores,   we calculate the similarity score \begin{math}
  s_{n}^{ocr} 
\end{math}   ($n=1, ..., N$) between each OCR token embedding  \begin{math}z_{n}^{ocr} \end{math} and the final answer representation,  where $N$ is the number of the OCR tokens. For fixed answer vocabulary scores,  the answer representation is fed into a classifier to output \begin{math}
  s_{m}^{voc} 
\end{math}   ($m=1, ..., M$) where $M$ is the size of the fixed vocabulary. 
This process is formulated as
\begin{align}
\begin{split}
     s_{n}^{ocr} &=   ( W^{ans}*z^{ans} + bias^{ans} )    ( W^{ocr}*z_{n}^{ocr} + bias^{ocr} )  ^{T}\\
    s_{m}^{voc} &= W^{voc}*z^{ans} + bias^{voc}
    \end{split}
\end{align}
where 
  $W^{ans}$, $W^{ocr}$, $W^{voc}$, $bias^{ans}$, $bias^{ocr}$, $bias^{voc}$
 are parameters of linear projection layers. The final prediction is the maximum score in the concatenation of vocabulary and OCR tokens 
  $max[s_{n}^{ocr}, s_{m}^{voc}]$.

There are two advantages of using this explicit semantic guidance. First, the predicted global semantic embedding works as an ``intuition'', which makes the decoder consider not only local decoding information but also global semantic information to generate more accurate results. Second, the pretrained FastText model provides supervision of external language knowledge, enabling the model to acquire the ability to distinguish the meanings of different words in the answer space.

 Furthermore, aiming to obtain a discriminative answer semantic embedding, our ICSP uses an instance-level contrastive method to push the predicted answer embedding closer to ground truth with others further away.  It 
 helps obtain a more semantically reasonable answer. As a result, the power of fixed vocabulary is fully exploited and the effect of OCR errors is alleviated.

\vspace{-1em}
\subsection{Training Loss}
\textbf{Binary Cross-Entropy Loss}
Considering that the answer may come from two sources  (fixed vocabulary and OCR tokens),   the multi-label binary cross-entropy  (BCE)  loss is adopted.
\begin{table*}[h]
 \centering
  \small
    \caption{Ablation study on the TextVQA dataset. M4C is our baseline method and the  uparrow in columns means  the performance improvement over M4C with same configurations. * indicates our ablations for improved baseline.  }
    
  \begin{tabular}{lllccll}
  
    \toprule
     \#&Model&OCR system &SCT & ICSP&Acc on val & Acc on test\\
  
    \midrule
    1&M4C (baseline) \cite{hu2020iterative} & Rosetta-en& -  &-&  39.40&  39.01\\
    2&SC-Net (ablation) & Rosetta-en&  \checkmark &-&  40.69  (1.29$\uparrow$) &  41.11  (2.10$\uparrow$) \\
    3&SC-Net (ablation) & Rosetta-en&  & \checkmark&  40.42  (1.02$\uparrow$) &  40.60  (1.59$\uparrow$) \\
    4&SC-Net  & Rosetta-en &  \checkmark & \checkmark & \textbf{41.17  (1.77$\uparrow$)} &  \textbf{41.42  (2.41$\uparrow$)} \\
  
     \hline
    5& M4C(baseline)* \cite{hu2020iterative} & SBD-Trans & - &- & 42.29 &  43.53  \\
    6&SC-Net & SBD-Trans& \checkmark &\checkmark&  \textbf{44.75  (2.46$\uparrow$)}&  \textbf{45.65 (2.12$\uparrow$)  } \\

    \bottomrule
  \end{tabular}

  \label{table1}
\end{table*}
\textbf{\\Instance-Level Contrastive Loss}
To get a better predicted semantic embedding of the answer,  we use the instance-level contrastive loss to evaluate the similarity of an answer's predicted semantic embedding \begin{math}
  pred_{semantic}
\end{math}  with all words' FastText embeddings  \begin{math} S
\end{math} in the fixed vocabulary and OCR tokens:
\begin{align}
    \begin{split}
        &L_{s} = -\sum_{i=1}^{K} log \frac{\sum_{j=1}^{M+N}exp ( S_{i, j}^{T}pred_{semantic}/\tau ) *y_{gt}}{\sum_{j=1}^{M+N}exp ( S_{i, j}^{T}pred_{semantic}/\tau ) * ( 1-y_{gt} ) } \\
        &L_{final} =L_{bce}+\alpha_{semantic} * L_{s}
         \end{split}
\end{align} 
The ground-truth label is  \begin{math}
  y_{gt}
\end{math}. \begin{math} K \end{math} is the batch size.  \begin{math} M \end{math} is the size of the fixed vocabulary and \begin{math} N \end{math} is the number of OCR tokens. $\tau$ is the temperature coefficient. 
\begin{math}
  \alpha_{semantic}
\end{math} is a hyper-parameter to control the trade-off between  \begin{math}
  L_{s}
 \end{math} and  \begin{math}
  L_{bce}
 \end{math}.

\section{EXPERIMENTS}
In this section,   we first 
compare our method with  baseline method  M4C   and other SOTA methods \cite{han2020finding,liu2020cascade,zhu2020simple}  on TextVQA. Extensive ablation experiments on TextVQA prove the effectiveness of our modules. Then we  compare our method with  SOTA methods on ST-VQA.  Implementation details and  visualization analysis detailed in supplementary material. 



  
  
    

 
\subsection{Evaluation on TextVQA Dataset}
The TextVQA dataset contains 28,408 images from the Open Images dataset with 45,336 questions, and each question has 10 answers. The final accuracy is measured by soft voting of 10 answers.\\

\textbf{Ablation Study on TextVQA Dataset}
To evaluate the effects of our proposed modules,  we perform an ablation experiment on the TextVQA dataset. Line 2 and Line 3 in Table \ref{table1} show our two modules SCT and ICSP improve the  performance over M4C on the test set by 2.10\% and 1.59\%, respectively. It demonstrates the effectiveness of each module. Especially,  SCT significantly improves the accuracy from 39.01\% to 41.11\%. It verifies that our SCT can better utilize the semantic information of the input and reinforce the multimodal representation.

Furthermore, integrating these modules with M4C leads to an accuracy improvement of 1.77\% and 2.41\% on the validation and test set respectively.
In order to compare with other methods using advanced OCR modules, we upgrade the OCR module to SBD-Trans \cite{liu2019omnidirectional} modules. With the same advanced OCR module, our SC-Net still achieve 2.46\% and 2.12\% improvements on the validation and test set.\\

\begin{table}[h]
 \centering
 \small
  \vspace{-1em}
  \caption{Comparison with existing methods on the TextVQA.}
  \setlength\tabcolsep{1pt}
  
  \begin{tabular}{llllll}
    \toprule
    \#&Model     &OCR system &Acc on val & Acc on test\\
  
    \midrule
    1&LoRRA \cite{singh2019towards}  &Rosetta-ml& 26.56&  27.63\\
    2&M4C \cite{hu2020iterative}   &Rosetta-en& 39.40&  39.01\\
    3&SMA \cite{ gao2020structured}  &Rosetta-en& 39.58&  40.29\\
    4&CRN \cite{liu2020cascade}  &Rosetta-en& 40.39&  40.96\\
    5&LaAP-Net \cite{han2020finding}   &Rosetta-en& 40.68&  40.54\\
    6&SC-Net   &Rosetta-en& \textbf{41.17}&  \textbf{41.42}\\
     \hline
    7&SSBaseline \cite{zhu2020simple}     &SBD-Trans& 43.95&  44.72\\
    8&SA-M4C  \cite{kant2020spatially}    &Google OCR& 43.90&  -\\
    9&SC-Net     &SBD-Trans& \textbf{44.75}&  \textbf{45.65}\\
   
    \bottomrule
  \end{tabular}
   
   \label{table2}
\end{table}

\textbf{Comparison with Existing Methods}
Table 2 compares our model and existing methods on TextVQA dataset. Equipped with Rosetta-en OCR module, LaAP-Net and CRN achieve the SOTA performance among previous methods on the val and test set, respectively. Especially, CRN utilizes the additional ANLS policy gradient loss which contributes 2.07\% absolute performance on test set. Compared with them, SC-Net surpasses LaAP-Net by 0.49\% on the val set and surpasses CRN by 0.46\% on the test set, proving the superiority of our method. It is important to note that M4C, SMA, and CRN take the object features as a part of the input into the transformer layer, while we use them to enhance the OCR token features.In this way, the input length reduce from 182 to 82, which saves a large amount of computation. In addition, with the same SBD-Trans OCR module, SC-Net surpasses the SSBaseline \cite{zhu2020simple} by 0.8\% and 0.93\% on the val and test set.

\subsection{Evaluation on ST-VQA Dataset}\label{4.3}
The ST-VQA dataset contains 18,921 training-validation images and 2,971 test images from multiple datasets  (ICDAR2013 ,  ICDAR2015 ,  ImageNet ,  VizWiz,  IIIT STR ,  Visual Genome ,  and COCO-Text ). Following previous works,  we use 17,028 images as the training set and 1,893 images as the validation set. The metric of ST-VQA dataset is Average Normalized Levenshtein Similarity   (ANLS),   defined as \begin{math}1-d_{L}   ( pred, gt )  /max  ( |pred|, |gt| )    \end{math},  where \begin{math}pred \end{math} is the prediction,  \begin{math} gt \end{math}  is the ground-truth answer, and  \begin{math} d_{L} \end{math} is the edit distance. Following CRN \cite{liu2020cascade}, we  adds the ANLS policy gradient loss in ST-VQA experiments.\

We report both accuracy and ANLS score in our experiments. As shown in Table \ref{table4},  our model surpasses the baseline M4C by a large margin. Specifically,  we improve 2.36\% in accuracy and 3.18\% in ANLS on the val set,  and 2.7\% in ANLS on the test set. Compared with other existing methods, our model is also superior to them in the both metrics.

\begin{table}[h]
  \centering
  \small
\vspace{-1em}
   \caption{Comparision on the ST-VQA dataset. }
   
   \scalebox{0.9}{\begin{tabular}{lllll}
  
    \toprule
     \#&Model&Acc on val & ANLS on val& ANLS on test\\

    \midrule
    1&VTA &-  & -&  0.282\\
    2&M4C\cite{hu2020iterative}& 38.05  & 0.4720&  0.462\\
    3&SMA\cite{ gao2020structured}  &-& -&  0.466\\
    4&CRN\cite{liu2020cascade}  &-& - & 0.483\\
    5&LaAP-Net\cite{han2020finding}   &39.74 & 0.4974&  0.485\\
    6&SC-Net  & \textbf{40.41}& \textbf{0.5038}&  \textbf{0.489}\\
    \bottomrule
  \end{tabular}}

  \vspace{-1em}
 
  \label{table4}
  
\end{table}

\section{CONCLUSION}
In this paper,  we propose a semantics-centered network to address several challenges in TextVQA task. First, considering the semantic information plays a critical role in TextVQA task, we design a novel semantics-centered transformer module  to emphasis the semantic information in the pipeline  and reduce large amount of computation.  Second, we propose a instance-level contrastive semantic prediction module  to provide a global semantic embedding of the answer and use it to guide the answer decoding. Following these strategies,  our model can locate the answer more accurately from a semantic perspective and meanwhile reduce the computational complexity. As a result, alleviating language bias and  OCR error problems. Extensive experiments demonstrate the proposed SC-Net outperforms previous works on TextVQA and ST-VQA datasets.

\bibliographystyle{IEEEbib}
\bibliography{icme2022template}

\begin{thebibliography}{10}

\bibitem{goyal2017making}
Yash Goyal, Tejas Khot, Douglas Summers-Stay, Dhruv Batra, and Devi Parikh,
\newblock ``Making the v in vqa matter: Elevating the role of image
  understanding in visual question answering,''
\newblock in {\em ICCV}, 2017, pp. 6904--6913.

\bibitem{anderson2018bottom}
Peter Anderson, Xiaodong He, Chris Buehler, Damien Teney, Mark Johnson, Stephen
  Gould, and Lei Zhang,
\newblock ``Bottom-up and top-down attention for image captioning and visual
  question answering,''
\newblock in {\em CVPR}, 2018.

\bibitem{jiang2018pythia}
Yu~Jiang, Vivek Natarajan, Xinlei Chen, Marcus Rohrbach, Dhruv Batra, and Devi
  Parikh,
\newblock ``Pythia v0. 1: the winning entry to the vqa challenge 2018,''
\newblock {\em CoRR}, vol. abs/1807.09956, 2018.

\bibitem{singh2019towards}
Amanpreet Singh, Vivek Natarajan, Meet Shah, Yu~Jiang, Xinlei Chen, Dhruv
  Batra, Devi Parikh, and Marcus Rohrbach,
\newblock ``Towards vqa models that can read,''
\newblock in {\em CVPR}, 2019, pp. 8317--8326.

\bibitem{hu2020iterative}
Ronghang Hu, Amanpreet Singh, Trevor Darrell, and Marcus Rohrbach,
\newblock ``Iterative answer prediction with pointer-augmented multimodal
  transformers for textvqa,''
\newblock in {\em CVPR}, 2020, pp. 9992--10002.

\bibitem{vaswani2017attention}
Ashish Vaswani, Noam Shazeer, Niki Parmar, Jakob Uszkoreit, Llion Jones,
  Aidan~N Gomez, Lukasz Kaiser, and Illia Polosukhin,
\newblock ``Attention is all you need,''
\newblock in {\em NeurIPS}, 2017, pp. 5998--6008.

\bibitem{liu2020cascade}
Fen Liu, Guanghui Xu, Qi~Wu, Qing Du, Wei Jia, and Mingkui Tan,
\newblock ``Cascade reasoning network for text-based visual question
  answering,''
\newblock in {\em ACM MM}, 2020.

\bibitem{han2020finding}
Wei Han, Hantao Huang, and Tao Han,
\newblock ``Finding the evidence: Localization-aware answer prediction for text
  visual question answering,''
\newblock in {\em COLING}, 2020.

\bibitem{zhu2020simple}
Qi~Zhu, Chenyu Gao, Peng Wang, and Qi~Wu,
\newblock ``Simple is not easy: A simple strong baseline for textvqa and
  textcaps,''
\newblock in {\em AAAI}, 2021.

\bibitem{yang2021tap}
Zhengyuan Yang, Yijuan Lu, Jianfeng Wang, Xi~Yin, Dinei Florencio, Lijuan Wang,
  Cha Zhang, Lei Zhang, and Jiebo Luo,
\newblock ``Tap: Text-aware pre-training for text-vqa and text-caption,''
\newblock in {\em CVPR}, 2021, pp. 8751--8761.

\bibitem{biten2019scene}
Ali~Furkan Biten, Ruben Tito, Andres Mafla, Lluis Gomez, Mar{\c{c}}al Rusinol,
  Ernest Valveny, CV~Jawahar, and Dimosthenis Karatzas,
\newblock ``Scene text visual question answering,''
\newblock in {\em ICCV}, 2019, pp. 4291--4301.

\bibitem{mishra2019ocr}
Anand Mishra, Shashank Shekhar, Ajeet~Kumar Singh, and Anirban Chakraborty,
\newblock ``Ocr-vqa: Visual question answering by reading text in images,''
\newblock in {\em ICDAR}, 2019.

\bibitem{wang2020general}
Xinyu Wang, Yuliang Liu, Chunhua Shen, Chun~Chet Ng, Canjie Luo, Lianwen Jin,
  Chee~Seng Chan, Anton van~den Hengel, and Liangwei Wang,
\newblock ``On the general value of evidence, and bilingual scene-text visual
  question answering,''
\newblock in {\em CVPR}, 2020, pp. 10126--10135.

\bibitem{gao2020structured}
Chenyu Gao, Qi~Zhu, Peng Wang, Hui Li, Yuliang Liu, Anton van~den Hengel, and
  Qi~Wu,
\newblock ``Structured multimodal attentions for textvqa,''
\newblock {\em CoRR}, vol. abs/2006.00753, 2020.

\bibitem{kant2020spatially}
Yash Kant, Dhruv Batra, Peter Anderson, Alex Schwing, Devi Parikh, Jiasen Lu,
  and Harsh Agrawal,
\newblock ``Spatially aware multimodal transformers for textvqa,''
\newblock in {\em ECCV}, 2020, pp. 715--732.

\bibitem{devlin2018bert}
Jacob Devlin, Ming-Wei Chang, Kenton Lee, and Kristina Toutanova,
\newblock ``Bert: Pre-training of deep bidirectional transformers for language
  understanding,''
\newblock in {\em NAACL}, 2019.

\bibitem{ren2015faster}
Shaoqing Ren, Kaiming He, Ross~B. Girshick, and Jian Sun,
\newblock ``Faster {R-CNN:} towards real-time object detection with region
  proposal networks,''
\newblock in {\em NeurIPS}, 2015.

\bibitem{bojanowski2017enriching}
Piotr Bojanowski, Edouard Grave, Armand Joulin, and Tomas Mikolov,
\newblock ``Enriching word vectors with subword information,''
\newblock {\em ACL}, pp. 135--146, 2017.

\bibitem{almazan2014word}
Jon Almaz{\'a}n, Albert Gordo, Alicia Forn{\'e}s, and Ernest Valveny,
\newblock ``Word spotting and recognition with embedded attributes,''
\newblock {\em IEEE TPAMI}, pp. 2552--2566, 2014.

\bibitem{liu2019omnidirectional}
Yuliang Liu, Sheng Zhang, Lianwen Jin, Lele Xie, Yaqiang Wu, and Zhepeng Wang,
\newblock ``Omnidirectional scene text detection with sequential-free box
  discretization,''
\newblock in {\em IJCAI}, 2019.

\end{thebibliography}
\appendix

\section{More Implementation Details}

In this section, we summarize the implementation and hyper-parameter settings of the  model designed in our work. All experiments are based on PyTorch deep-learning framework.

SC-Net consists of four main components: multimodal inputs, a multi-layer semantics-centered transformer, a multi-layer transformer and  instance-level contrastive semantic prediction. During training, the answer words are iteratively predicted and supervised using teacher-forcing. We apply multi-label sigmoid loss over the concatenation of ﬁxed answer vocabulary scores and dynamic OCR-copying scores.

For a fair comparison, most of settings in SC-Net are the same as in M4C. Specifically, the ﬁxed vocabulary in TextVQA dataset is composed of the top 5000 frequent words from the answers. We list the network and optimization parameters of SC-Net in Table \ref{table5} and Table \ref{table6} respectively.

\begin{table}[h]
  
  \caption{The network parameters of SC-Net. }
  \begin{tabular}{ll}
  
    \toprule
     Network parameters&Value\\\hline

    \midrule
   Maximum question tokens&{20}\\
    Maximum OCR tokens&{50}\\
    Maximum object tokens&{100}\\
    Text FastText embedding dimension&{300}\\
    Text PHOC embedding dimension&{604}\\
    Text Faster R-CNN feature dimension&{2048}\\
    Object Faster R-CNN feature dimension&{2048}\\
    Joint embedding dimension&{768}\\
    Semantics-Centered transformer layers&{2}\\
    Transformer layers&{2}\\
    Transformer attention heads&{12}\\
    Maximum decoding steps&{12}\\
    \bottomrule
  \end{tabular}
  
  \label{table5}
\end{table}

\begin{table}[h]
  
  \caption{The optimization parameters of SC-Net. }
  \begin{tabular}{ll}
  
    \toprule
     Optimization parameters&Value\\\hline

    \midrule
   Optimizer&{Adam}\\
    Base learning rate&{1e-4}\\
    Learning rate decay&{0.1}\\
    Warm-up iterations&{1000}\\
    Warm-up learning rate factor&{0.2}\\
    Batch size&{48}\\
    Max iterations&{48000}\\
    Learning rate decay steps&{28000, 38000}\\
    \bottomrule
  \end{tabular}
  
  \label{table6}
\end{table}

 \begin{figure*}[!htb]
  \centering
  \includegraphics[width=1\linewidth]{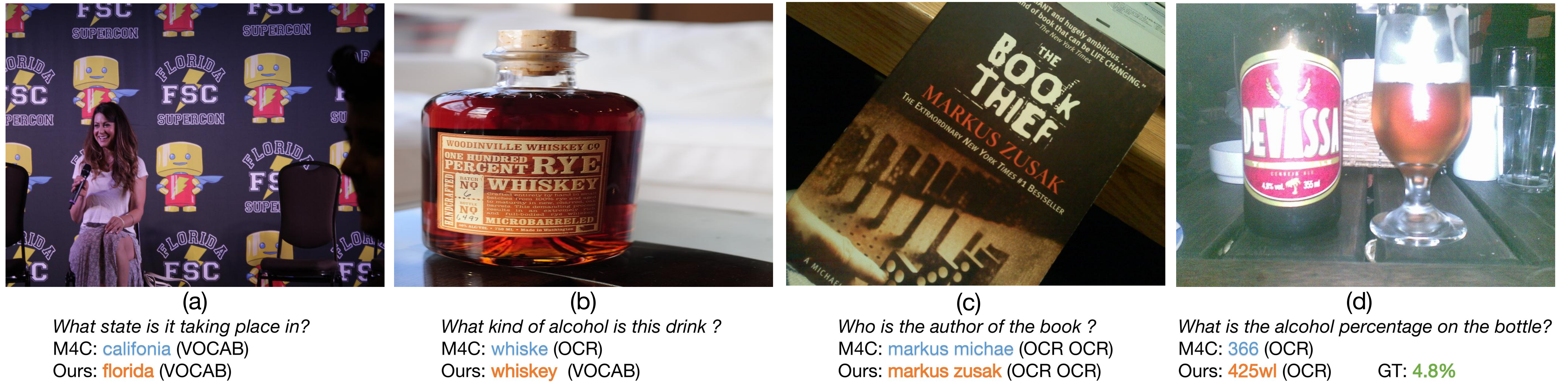}
  \caption{Qualitative examples of our model compared to the baseline M4C.}
  \label{figure_5}
\end{figure*}

\begin{figure*}[h]
    \centering
    \includegraphics[width=1\linewidth]{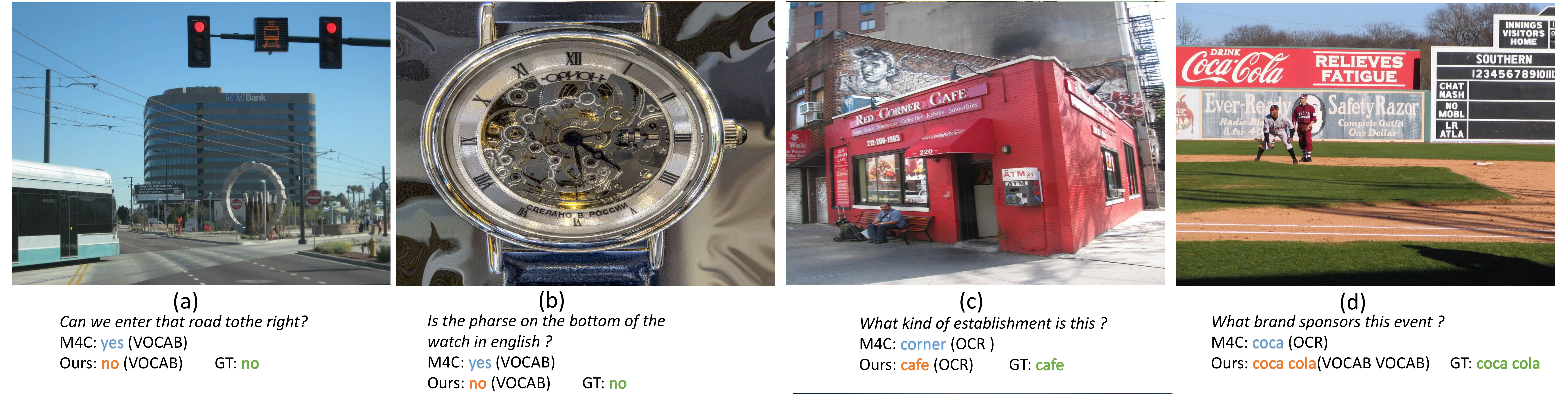}
     \includegraphics[width=1\linewidth]{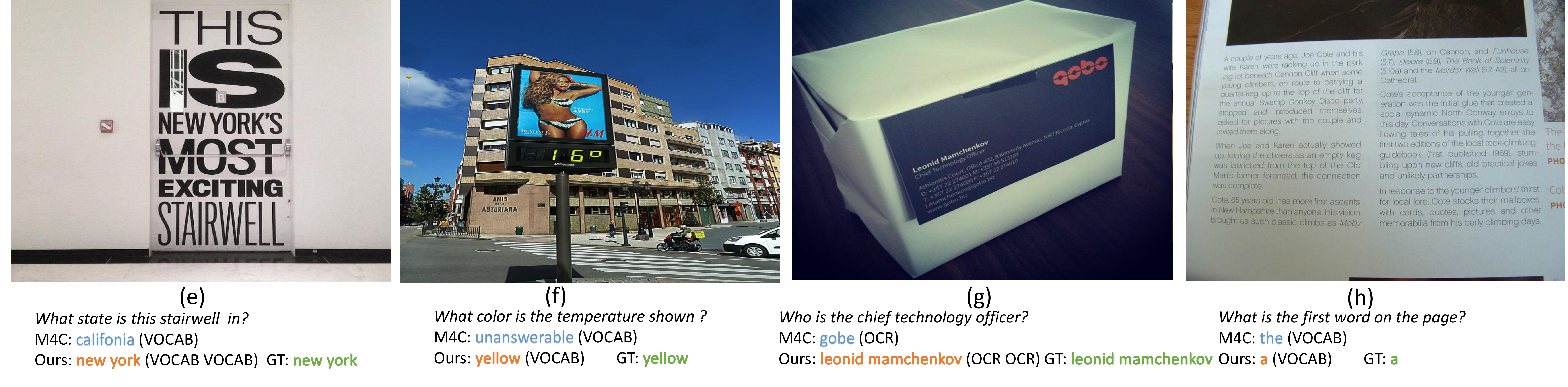}
    \caption{More qualitative examples on the TextVQA. The predictions of SC-Net (Ours) and M4C are depicted in orange and blue respectively. The ground-truth answer is depicted in green.}
    \label{fig:fig1}
\end{figure*}

\begin{table}[h]
  
  \caption{Ablation study with respect to answers' source of M4C and our model. Vocab means accuracy contributed by  vocabulary words. OCR means accuracy contributed by OCR words.}
  \begin{tabular}{lllll}
  \toprule
     \multirow{2}{*}{\shortstack{\#}} &  \multirow{2}{*}{\shortstack{Model}}  &  \multirow{2}{*}{\shortstack{Vocab}} & \multirow{2}{*}{\shortstack{OCR}} & \multirow{2}{*}{\shortstack{Acc     \\on val}} \\\\
    
    \midrule
    1&M4C  &13.48  & 25.92&  39.40\\
    2&M4C + ICSP &14.52  & 25.90&  40.42\\
    3&M4C + ICSP + SCT &15.15   & 26.02&  41.17\\

    \bottomrule
  \end{tabular}
  
  \label{table7}
\end{table}
\section{Ablation Study on Source of Answer}
To further explore the principle of the proposed methods, we design an ablation study  to analyze the performance improvements from different answer sources. If our method could alleviate language bias and OCR error problems, the model could distinguish low frequent words in the fixed vocabulary and choose the right text in place of error OCR.  As a result, the accuracy of the fixed vocabulary will increase.  As we can see in Table \ref{table7},
with the adoption of the ICSP module, the accuracy of answers from the fixed vocabulary improves by 1.04\%, while that from OCR results is almost unchanged. It indicates that introducing semantic information into the answer decoding module can better understand the
 meaning of words in the fixed vocabulary from a semantic perspective. It makes the model have a better discriminate ability to texts in vocabulary and use texts in vocabulary to compensate for OCR errors.
 
 In Line 3, equipped with the SCT module, the semantic information becoming the focus in the pipeline, the power of ICSP is further explored which benefits the accuracy from both fixed vocabulary and OCR tokens. The overall improvements show that it is reasonable and important to highlight semantic information in the multimodal fusion stage. It also demonstrates the complementary effect of ICSP and SCT.

\section{Qualitative Analysis}\label{4.4}
We show some visualization samples from TextVQA dataset in Figure \ref{figure_5} to validate the superiority of our model compared with the baseline M4C. With the guidance of semantic information, our model is able to take full advantages of the fixed vocabulary and modify many semantically irrational results. As shown in (a, b), the OCR module fails to recognize ``florida'' and  ``whisky'' as  ``elorida'' and ``whiske'' respectively. For (a), M4C is misled by the wrong OCR result and chooses the word ``california'' in the fixed vocabulary, which frequently occurs together with the question word ``state''. For (b), M4C is not aware of the recognition error, simply copying the OCR result as the answer. On the contrary, SC-Net considers the global semantics of question, image features and text features comprehensively, locating the accurate answer from the fixed vocabulary. It proves that our method can effectively conquer the data biases and make better use of semantic information conductive to answer prediction. As shown in (c), the design of SCT module enables SC-Net to associate the visual information of texts (the red color of ``markus'' and ``zusak'') and find related words to compose a more semantically appropriate answer. One failure case is shown in (d),  the ground truth answer of the question is ``4.8\%'' whereas the M4C gives the irrelevant OCR token ``366'' and our model predicts ``425wl'' as the answer. It is interpretable why our method does not answer the correct number based on semantic information. Even humans have difficulty providing correct answers from fuzzy numbers and the correct answer “4.8\%” is not included in the fixed vocabulary. However,  our model still finds the correct answer region and gives the word most similar to ``4.8\%'' in semantics.

\section{More Visualization Examples}

We provide more qualitative examples in Figure \ref{fig:fig1}  to support our claim in the paper.

\begin{figure*}[h]
    \centering
    \includegraphics[width=1\linewidth]{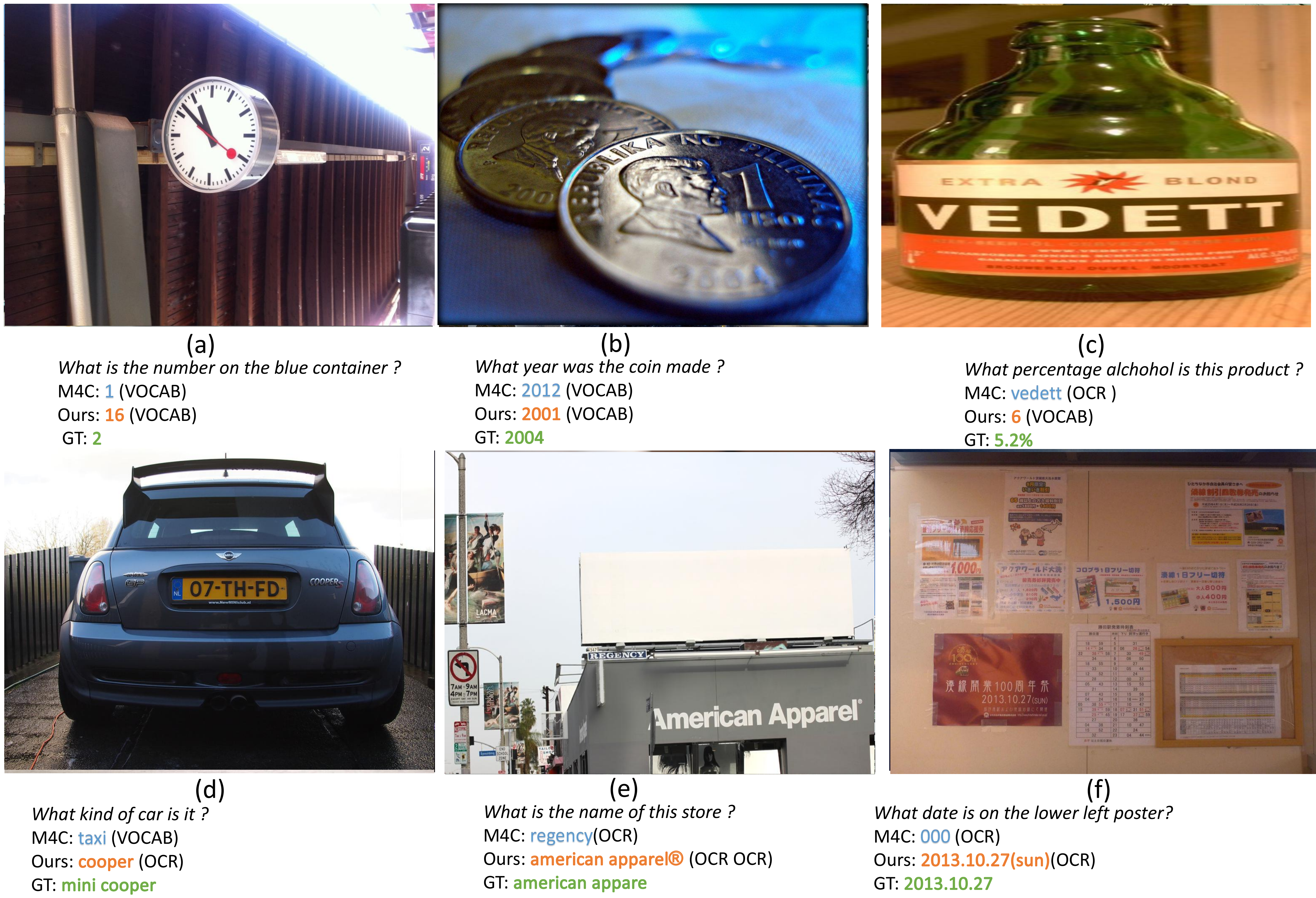}
    \caption{Some failure cases of SC-Net (Ours) model on the TextVQA validation set. The predictions of SC-Net (Ours) and M4C are depicted in orange and blue respectively. The ground-truth answer is depicted in green.}
    \label{fig:fig2}
\end{figure*}

There are still some failure cases (Figure \ref{fig:fig2}) in our experiments, which are extremely challenging. 

Firstly, the time of the clock and watch is difficult to recognized correctly by the OCR system. In (a), M4C and our SC-Net model cannot predict the right answer in this situation.  

Secondly, since there are no specific year numbers and alcohol levels in the dictionary, our ICSP module cannot select the correct answer from the fixed vocabulary by answer's semantic information. As shown in (b) and (c), our model can still  answer with more reasonable words than M4C.

In (d), our SC-Net model answer with ``cooper'' in OCR tokens, which is closer to the answer ``mini cooper'' than ``taxi'' in semantics. In addition, we find some interesting failure cases, in (e) and (f), our model answer with  more accurate answers that ground truth answers.  However, the accuracy calculation still treats them as wrong answers.

\end{document}